\documentclass[a4paper,twocolumn]{article}
\usepackage{rsj2025e}
\usepackage[dvipdfmx]{graphicx}
\usepackage{amsmath}
\usepackage{url}
\usepackage{inconsolata}
\usepackage{graphicx}
\usepackage{amsmath}
\usepackage{amssymb}
\usepackage{booktabs}
\usepackage{amsthm}  
\usepackage{amsmath,amssymb,amsfonts}
\usepackage{multirow}
\usepackage{graphicx} 
\usepackage{booktabs}
\usepackage{textcomp}
\usepackage{xcolor}
\usepackage{enumitem}
\usepackage{algorithm}
\usepackage{algpseudocode}
\usepackage{amsmath}
\usepackage{url} 
\usepackage{multirow}
\usepackage{graphicx}
\usepackage{tabularx}
\usepackage{booktabs}
\usepackage{cite}
\usepackage{subcaption}
\usepackage{caption}
\usepackage{tikz}
\usepackage{amsmath}
\usepackage{amssymb}
\usetikzlibrary{shapes.geometric}
\usetikzlibrary{shapes}
\usetikzlibrary{arrows,decorations.pathmorphing,backgrounds,positioning,fit,petri,shadows,calc,arrows.meta}
\begin{document}
\title{Knowledge-Augmented Vision Language Models for Underwater Bioacoustic Spectrogram Analysis}
\author{
Ragib Amin Nihal$^{*1}$, 
  Benjamin Yen$^{1,2}$, 
  Takeshi Ashizawa$^{1}$, 
  Kazuhiro Nakadai$^{1}$\\
  \small{$^1$Systems and Control Engineering, Institute of Science Tokyo}\\
  \small{$^2$RIKEN}
}

\noindent\abstract{Marine mammal vocalization analysis depends on interpreting bioacoustic spectrograms. Vision Language Models (VLMs) are not trained on these domain-specific visualizations. We investigate whether VLMs can extract meaningful patterns from spectrograms visually. Our framework integrates VLM interpretation with LLM-based validation to build domain knowledge. This enables adaptation to acoustic data without manual annotation or model retraining.}

\setlength{\baselineskip}{4.4mm}
\maketitle
\thispagestyle{empty}
\pagestyle{empty}
\let\thefootnote\relax\footnotetext{$^*$Corresponding author: Ragib Amin Nihal (\texttt{ragib@ra.sc.e.titech.ac.jp})}
\section{Introduction}
Marine mammals depend on acoustic communication for navigation, social interactions, and finding food across vast ocean environments. As climate change and human activities threaten many species with extinction, understanding these vocalizations has become essential for conservation efforts under Sustainable Development Goal 14 -- \textit{Life Below Water}.
\\
Automatically classifying marine mammal sounds from recordings presents major challenges. Underwater soundscapes are complex, each species has unique vocal patterns, and interpreting acoustic features requires biological expertise.
\\
Current approaches face a three-way trade-off between performance, cost, and interpretability. Specialized CNN architectures like MG-ResFormer and WhaleNet achieve 97-99\% accuracy \cite{liu2024classification,zheng2024mt,schall2024deep} but operate as black boxes, providing no biological insight for expert validation \cite{shiu2020deep}. When researchers encounter new species or different acoustic environments, these systems require expensive retraining, large annotated datasets, and specialized expertise. Even the recent NatureLM-audio foundation model, while promising, still needs fine-tuning and struggles with unseen species \cite{hagge2024naturelm}.
\\
Conservation biologists need systems that can explain their decisions, work with new species without retraining, and integrate human expertise. This creates a fundamental challenge: specialized accuracy versus interpretable adaptability. 
Vision Language Models (VLMs) \cite{radford2021learning} offer a potential solution. These models can generate natural language descriptions of acoustic patterns, much like they do for medical imaging \cite{johnsnow2024medical}. Recent work shows VLMs can analyze spectrograms \cite{gardner2024vision}, but models trained on general internet data lack the specialized biological knowledge needed for accurate bioacoustic analysis.
\begin{figure}[tb]
\centering
\begin{tikzpicture}[
    scale=0.45,                           
    every node/.style={scale=0.45},       
    node distance=1.5cm                   
]

\tikzstyle{audio box} = [draw, rectangle, fill=blue!10, minimum width=2.5cm, minimum height=1cm]      
\tikzstyle{spec box} = [draw, rectangle, fill=green!10, minimum width=2.5cm, minimum height=1cm]     
\tikzstyle{vlm box} = [draw, rectangle, fill=red!20, minimum width=3.2cm, minimum height=1.3cm]      
\tikzstyle{output box} = [draw, rectangle, fill=cyan!20, minimum width=2.5cm, minimum height=1cm]    
\tikzstyle{kb box} = [draw, rectangle, fill=purple!15, minimum width=2.8cm, minimum height=1.2cm]    
\tikzstyle{thick arrow} = [->, thick, line width=0.5pt]                                              

\node[above] at (6,13) {\textbf{Vanilla VLM (Baseline)}};
\node[audio box] (audio1) at (0,12) {\shortstack{Audio\\Waveform}};
\node[spec box] (spec1) at (3.5,12) {\shortstack{Spectrogram}};
\node[vlm box] (vlm1) at (7.5,12) {\shortstack{Vision Language\\Model (VLM)}};
\node[output box] (output1) at (12,12) {\shortstack{Species\\Prediction}};

\draw[thick arrow] (audio1) -- (spec1);
\draw[thick arrow] (spec1) -- (vlm1);
\draw[thick arrow] (vlm1) -- (output1);

\draw[gray, thick, line width=0.8pt] (-1,10.8) -- (14,10.8);

\node[above] at (6,9.5) {\textbf{Fixed Knowledge Base}};
\node[audio box] (audio2) at (0,8.5) {\shortstack{Audio\\Waveform}};
\node[spec box] (spec2) at (3.5,8.5) {\shortstack{Spectrogram}};
\node[vlm box] (vlm2) at (7.5,8.5) {\shortstack{Vision Language\\Model (VLM)}};
\node[output box] (output2) at (12,8.5) {\shortstack{Species\\Prediction}};
\node[kb box] (kb_fixed) at (7.5,6.5) {\shortstack{Fixed\\Knowledge Base}};

\draw[thick arrow] (audio2) -- (spec2);
\draw[thick arrow] (spec2) -- (vlm2);
\draw[thick arrow] (vlm2) -- (output2);
\draw[thick arrow, dashed] (kb_fixed) -- (vlm2);

\draw[gray, thick, line width=0.8pt] (-1,5.8) -- (14,5.8);

\node[above] at (6,5) {\textbf{Progressive Knowledge Base (Proposed)}};

\node[draw, rectangle, fill=red!10, minimum width=2.5cm, minimum height=0.8cm, font=\bf] at (-2.5,4) {\shortstack{Learning\\Stage}};

\node[audio box] (audio3a) at (0,4) {\shortstack{Training\\Audio}};
\node[spec box] (spec3a) at (3,4) {\shortstack{Spectrogram}};
\node[vlm box] (vlm3a) at (6.5,4) {\shortstack{VLM Pattern\\Extraction}};
\node[draw, diamond, fill=yellow!20, minimum width=1.8cm, minimum height=1.5cm] (filter) at (10,4) {\shortstack{Quality\\Filter}};
\node[kb box] (kb_prog) at (13,4) {\shortstack{Progressive\\Knowledge Base}};

\node[draw, rectangle, fill=red!10, minimum width=2.5cm, minimum height=0.8cm, font=\bf] at (-2.5,1.5) {\shortstack{Inference\\Stage}};

\node[audio box] (audio3b) at (0,1.5) {\shortstack{Test\\Audio}};
\node[spec box] (spec3b) at (3,1.5) {\shortstack{Spectrogram}};
\node[vlm box] (vlm3b) at (6.5,1.5) {\shortstack{VLM Pattern\\Extraction}};
\node[draw, rectangle, fill=orange!20, minimum width=2.8cm, minimum height=1.2cm] (similarity) at (10,1.5) {\shortstack{NLP Similarity\\Engine}};
\node[output box] (output3) at (13,1.5) {\shortstack{Species\\Classification}};

\draw[thick arrow] (audio3a) -- (spec3a);
\draw[thick arrow] (spec3a) -- (vlm3a);
\draw[thick arrow] (vlm3a) -- (filter);
\draw[thick arrow] (filter) -- (kb_prog);

\draw[thick arrow] (audio3b) -- (spec3b);
\draw[thick arrow] (spec3b) -- (vlm3b);
\draw[thick arrow] (vlm3b) -- (similarity);
\draw[thick arrow] (similarity) -- (output3);
\draw[thick arrow, dashed] (kb_prog) -- (similarity);

\end{tikzpicture}
\caption{\scriptsize{Comparison of three marine mammal classification approaches. \textbf{Top:} Vanilla VLM approach processes spectrograms directly without domain knowledge, achieving baseline performance. \textbf{Middle:} Fixed Knowledge Base approach augments VLM with static expert-generated patterns for each species, providing consistent domain knowledge. \textbf{Bottom:} Progressive Knowledge Base approach (proposed) combines synthetic expert knowledge with AI-learned patterns through iterative learning stages, dynamically expanding the knowledge base through quality-filtered pattern extraction and similarity-based classification.}}
\label{fig:system_comparison}
\vspace*{-0.5cm}
\end{figure}
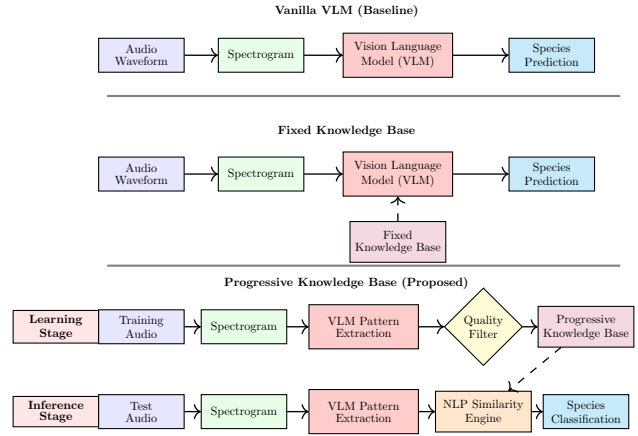
\\
We hypothesize that adding progressive knowledge accumulation to VLMs can bridge this gap while keeping costs low and maintaining interpretability. Knowledge augmentation through Retrieval-Augmented Generation has proven effective for enhancing models without expensive retraining \cite{lewis2020retrieval,yang2024crag}. Unlike methods like LoRA that modify the model itself \cite{hu2021lora}, knowledge augmentation works by adding external information.
\\
\textbf{Research Question}: Can knowledge-augmented VLMs achieve meaningful marine mammal classification performance while eliminating retraining requirements and providing interpretable pattern descriptions? \textbf{Central Hypothesis}: Providing sufficient contextual knowledge through progressive pattern accumulation can bridge the semantic gap between general VLM capabilities and domain-specific bioacoustic requirements. 
\\
In this research, we investigate whether progressive knowledge accumulation can bridge the semantic gap between general VLM capabilities and domain-specific bioacoustic requirements while maintaining the interpretability essential for expert validation. We reframe marine mammal classification as a two-stage process that separates pattern recognition from species identification using natural language descriptions.
\\
\textbf{Work in Progress Contributions:} This preliminary study demonstrates: (1) feasibility of interpretable pattern extraction from marine mammal spectrograms using VLMs, generating biologically meaningful descriptions like ``vertical burst patterns at 2-8 kHz with 50ms intervals"; (2) progressive knowledge accumulation that improves classification performance without model retraining, achieving 92\% improvement over vanilla VLMs; and (3) practical workflows for expert-assisted bioacoustic annotation where interpretability enables rapid species screening and validation.
\\
Our current accuracy (25.4\% across 31 species) falls short of specialized models, but we show feasibility for conservation applications that prioritize interpretability and rapid deployment over maximum autonomous performance. This work provides a foundation for human-AI collaborative bioacoustic monitoring where natural language pattern descriptions enable expert validation and iterative knowledge refinement.

\section{Problem Formulation}

\subsection{Mathematical Problem Statement}

Let $S = \{s_1, s_2, ..., s_n\}$ be a set of marine mammal species, and $X$ be the space of spectrogram images. Traditional supervised learning seeks to learn a mapping $f: X \rightarrow S$ using labeled training data $D = \{(x_i, y_i)\}_{i=1}^N$ where $x_i \in X$ and $y_i \in S$.
\\
\textbf{The Bioacoustic Classification Trilemma:} Existing approaches face the challenge of optimizing performance, cost-effectiveness, and interpretability:
$\max_{f} P(y|x)$
subject to constraints on retraining costs, annotation requirements, and interpretability needs. Traditional approaches require expensive retraining for each new species or environment.

\subsection{Knowledge-Augmented Formulation}

\noindent Our approach reformulates as a two-stages:
\\
\textbf{Stage 1-Pattern Extraction:}\\
$p_i = \text{VLM}(x_i, \text{prompt}_{\text{extract}})$
where $p_i$ is a natural language description of acoustic patterns in spectrogram $x_i$.
\\
\textbf{Stage 2-Knowledge-Augmented Classification:}\\
$$\hat{y} = \arg\max_{s \in S} \{\text{similarity}(p_i, KB(s))\}$$
where $KB(s)$ represents accumulated knowledge patterns for species $s$.

\subsection{Progressive Knowledge Base Evolution}

The knowledge base evolves iteratively:
$$KB^{(t+1)}(s) = KB^{(t)}(s) \cup \{p_j : Q(p_j) > \theta \land N(p_j) > \delta\}$$
where $Q(p_j)$ measures pattern quality, $N(p_j) = N(p_j, KB^{(t)}(s))$ measures novelty relative to existing knowledge, $\theta$ is a quality threshold, and $\delta$ ensures pattern diversity.
\begin{figure}[tb]
\centering
\begin{tikzpicture}[
    scale=0.4,                            
    every node/.style={scale=0.4},        
    node distance=0.8cm                   
]

\tikzstyle{sample box} = [draw, ellipse, fill=blue!10, minimum width=2cm, minimum height=1.2cm]        
\tikzstyle{pattern box} = [draw, rectangle, fill=yellow!10, minimum width=2.5cm, minimum height=1.2cm] 
\tikzstyle{process box} = [draw, rectangle, fill=red!20, minimum width=3cm, minimum height=3.5cm]      
\tikzstyle{kb box large} = [draw, rectangle, fill=purple!15, minimum width=3.5cm, minimum height=3cm]  

\node[sample box] (s1) at (0,6.5) {\shortstack{Sample 1\\(Species A)}};
\node[sample box, fill=blue!10] (s2) at (0,5) {\shortstack{Sample 2\\(Species A)}};
\node[sample box, fill=green!10] (s3) at (0,3.5) {\shortstack{Sample 3\\(Species B)}};
\node[sample box, fill=green!10] (s4) at (0,2) {\shortstack{Sample 4\\(Species B)}};

\node[process box] (vlm) at (4.5,4.25) {\shortstack{VLM\\Pattern\\Extraction}};

\node[pattern box] (p1) at (8.5,6.5) {\shortstack{Pattern P1\\``Vertical lines''}};
\node[pattern box] (p2) at (8.5,5) {\shortstack{Pattern P2\\``Burst sequences''}};
\node[pattern box] (p3) at (8.5,3.5) {\shortstack{Pattern P3\\``Harmonic sweeps''}};
\node[pattern box] (p4) at (8.5,2) {\shortstack{Pattern P4\\``Rhythmic calls''}};

\node[draw, diamond, fill=orange!20, minimum width=2cm, minimum height=2cm] (filter) at (12.5,4.25) {\shortstack{Quality \&\\Novelty\\Filter}};

\node[kb box large] (kb) at (16.5,4.25) {\shortstack{Progressive\\Knowledge Base\\\\Species A:\\P1, P2\\\\Species B:\\P3}};

\node[draw, rectangle, fill=red!20, minimum width=2cm, minimum height=1cm] (rejected) at (12.5,7) {Rejected P4};

\draw[->, thick, line width=0.5pt] (s1) -- (vlm);
\draw[->, thick, line width=0.5pt] (s2) -- (vlm);
\draw[->, thick, line width=0.5pt] (s3) -- (vlm);
\draw[->, thick, line width=0.5pt] (s4) -- (vlm);

\draw[->, thick, line width=0.5pt] (vlm) -- (p1);
\draw[->, thick, line width=0.5pt] (vlm) -- (p2);
\draw[->, thick, line width=0.5pt] (vlm) -- (p3);
\draw[->, thick, line width=0.5pt] (vlm) -- (p4);

\draw[->, thick, line width=0.5pt] (p1) -- (filter);
\draw[->, thick, line width=0.5pt] (p2) -- (filter);
\draw[->, thick, line width=0.5pt] (p3) -- (filter);
\draw[->, thick, line width=0.5pt] (p4) -- (filter);

\draw[->, thick, line width=0.5pt] (filter) -- node[above, rotate=8, font=\small] {accept} (kb);
\draw[->, thick, line width=0.5pt] (filter) -- node[left, font=\small] {reject} (rejected);

\draw[->, thick, dashed, red, line width=2pt] (kb) to[bend left=60] node[above, font=\small] {Next Iteration} (vlm);

\end{tikzpicture}
\caption{\scriptsize{Progressive learning algorithm showing pattern extraction, filtering, and knowledge base growth}}
\label{fig:progressive_learning}
\vspace{-.4cm}
\end{figure}
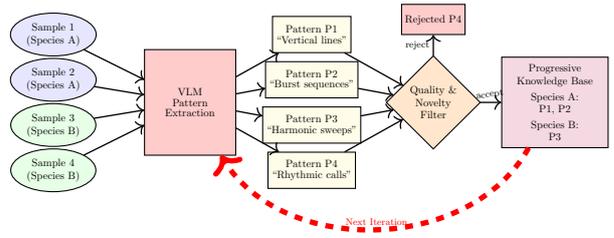
\subsection{Central Hypothesis}
We hypothesize that providing sufficient contextual knowledge through progressive pattern accumulation can bridge the semantic gap between general VLM capabilities and domain-specific bioacoustic requirements. Following standard statistical hypothesis testing, we define:
$$H_0: \text{acc}(\text{VLM} + \text{Context}) \leq \text{acc}(\text{VLM}_{\text{vanilla}})$$
$$H_1: \text{acc}(\text{VLM} + \text{Context}) > \text{acc}(\text{VLM}_{\text{vanilla}}),$$
where $H_0$ represents the null hypothesis (no improvement from context) that we aim to reject, and $H_1$ represents our alternative hypothesis (context provides improvement) that we aim to support. Additionally, we expect:
$$\text{cost}(\text{VLM} + \text{Context}) \ll \text{cost}(\text{retraining})$$


\section{Methodology}


\subsection{System Overview}

Our framework consists of three main components: (1) Pattern Extractor that uses VLMs to describe acoustic patterns, (2) Progressive Knowledge Base for dynamic pattern storage and retrieval, and (3) NLP Similarity Matcher using TF-IDF based pattern matching for classification. Our experimental design compares three distinct approaches to VLM-based marine mammal classification:
\\
\textbf{Vanilla VLM System:} Direct spectrogram classification using general-purpose VLMs without domain knowledge. This baseline establishes the performance floor for unaugmented systems.
\\
\textbf{Fixed Knowledge Base System:} VLMs augmented with static, internet collected acoustic pattern descriptions for each species. Knowledge remains constant throughout classification, providing consistent domain expertise.
\\
\textbf{Progressive Knowledge Base System (Proposed):} Dynamic system combining Fixed knowledge with AI-learned patterns. The knowledge base grows iteratively through quality-filtered pattern extraction from learning stage examples.
Figures \ref{fig:system_comparison} and \ref{fig:progressive_learning} illustrate these approaches and our learning algorithm.



\begin{algorithm}[tb]
\caption{Progressive Knowledge Accumulation}
\label{alg:progressive_learning}
\begin{algorithmic}[1]
\Require Training samples $\{(x_i, y_i)\}_{i=1}^N$, iterations $T$, samples per iteration $K$
\Ensure Enhanced knowledge base $KB$

\State Initialize $KB$ with synthetic expert patterns
\For{$t = 1$ to $T$}
    \For{each species $s$ in random sample}
        \State Sample $K$ examples from training set of species $s$
        \For{each example $x$}
            \State $p \leftarrow$ VLM\_extract\_pattern($x$)
            \If{quality($p$) $> \theta_q$ \textbf{and} novelty($p$, $KB(s)$) $> \theta_n$}
                \State $KB(s) \leftarrow KB(s) \cup \{p\}$
            \EndIf
        \EndFor
    \EndFor
    \State Update similarity matcher with enhanced $KB$
\EndFor
\State \Return $KB$
\end{algorithmic}
\end{algorithm}
\subsection{Progressive Learning Algorithm}

Algorithm \ref{alg:progressive_learning} implements standard progressive learning: for each iteration, sample $K$ examples per species, extract patterns via VLM, filter by quality and novelty thresholds, and update the knowledge base.
\subsection{NLP-Based Similarity Matching}

\noindent\textbf{TF-IDF Vectorization:} Patterns undergo TF-IDF vectorization with n-gram range (1,3) for phrase capture and domain-specific stop word filtering.

\noindent\textbf{Pattern-Level Similarity:}
$\text{sim}(p_{\text{query}}, p_{\text{kb}}) = \frac{v_{\text{query}} \cdot v_{\text{kb}}}{||v_{\text{query}}|| \cdot ||v_{\text{kb}}||}$
\\
\textbf{Species-Level Aggregation:}
\begin{align*}
\text{score}(s) &= 0.6 \cdot \max_{p \in KB(s)} \text{sim}(p_{\text{query}}, p) \nonumber\\
&\quad + 0.3 \cdot \text{mean}_{p \in KB(s)} \text{sim}(p_{\text{query}}, p) \nonumber\\
&\quad + 0.1 \cdot \text{diversity}(s)
\end{align*}

Final classification selects the species with maximum aggregated score: $\hat{y} = \arg\max_{s \in S} \text{score}(s)$.

\section{Experiments, Results and Discussion}

\subsection{Experimental Setup}

We used the Watkins Marine Mammal Sound Database \cite{sayigh2016watkins,watkins1993marine}. We tested different scenarios: 5, 10, and 24 species with 1, 3, and 5 samples per learning round. We ran experiments on both Qwen2.5-VL-7B and 32B models, using proper time-based validation across 60 total experiments.
\\

\subsection{Performance Results}

\noindent\textbf{Overall Performance:} From Table 1, we can see that, our progressive knowledge-augmented system achieved 25.4\% ± 11.5\% accuracy across all configurations, representing a 92\% improvement over vanilla VLMs (13.2\% baseline).


\noindent\textbf{Learning Dynamics:} Progressive knowledge base accumulated 51.3 ± 41.1 patterns per experiment, growing from fixed baseline (28.6 patterns) to enhanced knowledge (79.8 patterns). Table 2 shows performance vs. species complexity. 

\subsection{Quality of Generated Patterns}

The VLM generated biologically meaningful descriptions. For example:\\
- \textit{Humpback Whale}: ``Complex melodic sequences sweeping 20 Hz to 4 kHz with repetitive phrase structures"\\
- \textit{Bottlenose Dolphin}: ``High-frequency signature whistles with unique contour patterns around 8-12 kHz"\\  
- \textit{Fin Whale}: ``Powerful low-frequency pulse trains at 20 Hz with regular temporal intervals"\\
However, 23\% of patterns were too generic (``rhythmic calling behavior"), which cost classification performance. As expected, accuracy dropped with more species to distinguish between (Table \ref{tab:nway_analysis}).

\begin{table}[tb]
\centering
\caption{\scriptsize{System Performance and Feature Comparison}}
\resizebox{\columnwidth}{!}{%
\begin{tabular}{lcccccc}
\toprule
\textbf{Approach} & \textbf{Accuracy} & \textbf{No Retrain} & \textbf{Interpret.} & \textbf{Expert Int.} & \textbf{Cross-Spec.} & \textbf{Sci. Valid.} \\
\midrule
Traditional CNN & 97-99\% & $\times$ & $\times$ & $\times$ & $\times$ & $\times$ \\
NatureLM-audio (seen) & 75\% & $\times$ & \checkmark & \checkmark & $\times$ & \checkmark \\
NatureLM-audio (unseen) & 20\% & $\times$ & \checkmark & \checkmark & $\times$ & \checkmark \\
Vanilla VLM & 12.5\% & \checkmark & \checkmark & $\times$ & \checkmark & $\times$ \\
Fixed Knowledge Base & 18.5\% & \checkmark & \checkmark & \checkmark & \checkmark & \checkmark \\
Our Progressive VLM & 25.4\% & \checkmark & \checkmark & \checkmark & \checkmark & \checkmark \\
\bottomrule
\end{tabular}%
}
\label{tab:performance_comparison}
\end{table}


\begin{table}[tb]
\centering
\caption{\scriptsize{Performance vs. Species Complexity}}
\resizebox{\columnwidth}{!}{%
\begin{tabular}{lcccc}
\toprule
\textbf{N-way} & \textbf{Baseline Acc.} & \textbf{Progressive Acc.} & \textbf{Improvement} & \textbf{Patterns} \\
\midrule
5 species & 20.3\% & 31.8\% ± 13.3\% & +84.7\% & 28.6 ± 13.4 \\
10 species & 10.5\% & 25.4\% ± 8.1\% & +143.6\% & 51.2 ± 20.1 \\
24 species & 4.3\% & 16.2\% ± 2.4\% & +279.4\% & 79.8 ± 53.5 \\
\bottomrule
\end{tabular}%
}
\label{tab:nway_analysis}
\vspace{-.5cm}
\end{table}
\subsection{Discussion}



\begin{figure}[tb]
\centering
\includegraphics[width=1.05\columnwidth]{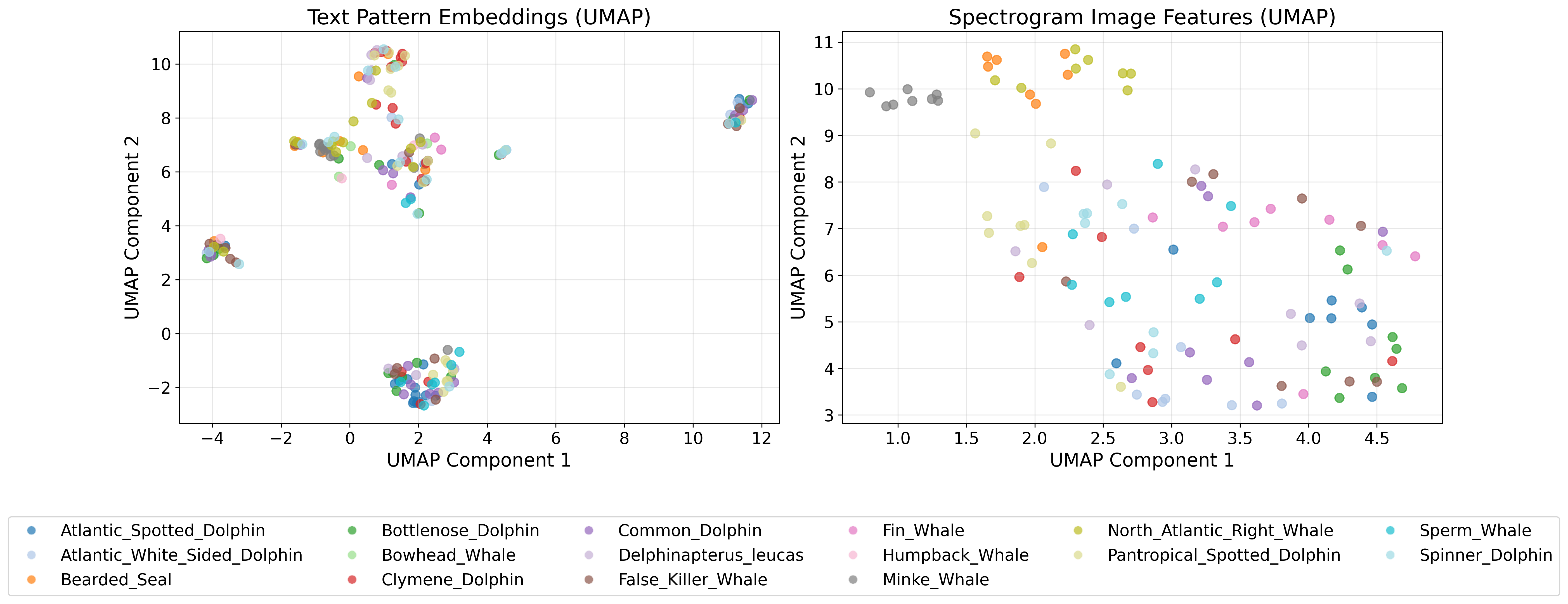}
\caption{\scriptsize{UMAP analysis shows semantic gap between visual features and textual pattern descriptions.}}
\label{fig:umap_analysis}
\vspace*{-0.5cm}
\end{figure}

\noindent\textbf{Interpretability Benefits:} Generated natural language pattern descriptions enable expert validation and biological insight extraction, addressing critical interpretability gaps in black-box CNN approaches. This facilitates human-AI collaboration for knowledge refinement.
\\
\textbf{Expert Validation Workflow Example:} When the system classifies a call as Fin Whale based on ``20 Hz pulse pattern with 12-second intervals," biologists can immediately assess: (1) frequency accuracy (Fin whales typically vocalize at 15-25 Hz), (2) temporal pattern validity (intervals of 8-15 seconds are characteristic), and (3) acoustic context appropriateness. This interpretability enables rapid validation decisions are not capable with black-box CNN outputs.
\\
\textbf{Semantic Gap Analysis:} UMAP \cite{mcinnes2018umap} visualization (Figure \ref{fig:umap_analysis}) reveals that VLM-generated pattern descriptions cluster by linguistic similarity rather than biological relationships, explaining performance limitations. Patterns like ``burst sequences" and ``rhythmic calls" appear semantically similar to the VLM despite representing different species, highlighting the challenge of bridging visual-acoustic features to meaningful biological descriptions. This semantic clustering problem suggests that improved prompt engineering and domain-specific fine-tuning could significantly enhance pattern discriminative power.
\\
\textbf{Quality-Quantity Trade-off Analysis:} The negative correlation between pattern quantity and accuracy (r=-0.44) shows that as the knowledge base grows, accumulated low-quality patterns introduce noise that degrades classification performance. Generic patterns like ``Rhythmic calling behavior" match multiple species indiscriminately, while specific patterns like ``20 Hz pulse trains with 12-second intervals" provide discriminative power. This suggests that pattern curation and quality thresholding are more critical than knowledge base size for improving accuracy.
\\
\textbf{Practical Applications:} Our system best serves as a rapid screening tool for preliminary species detection or as a bootstrap for annotation efforts, rather than a replacement for high-accuracy specialized models in critical conservation monitoring.
\\
\section{Conclusion and Future Directions}

This preliminary study demonstrates the feasibility of interpretable VLM-based marine mammal classification through progressive knowledge accumulation. While our current accuracy (25.4\%) cannot replace specialized models, we establish a foundation for applications where interpretability and rapid deployment outweigh maximum performance requirements.
Our approach uniquely provides natural language pattern descriptions enabling expert validation, eliminates retraining requirements for new species, and supports human-AI collaborative workflows essential for conservation monitoring. Generated patterns like ``vertical burst sequences at 2-8 kHz" offer biological insight absent from black-box approaches.
\\
\textbf{Ongoing Work:} We are developing enhanced prompt engineering strategies, hierarchical knowledge organization, and feedback integration to improve both accuracy and interpretability. Future evaluation will focus on (1) enhanced visual-linguistic alignment through contrastive learning, (2) hierarchical knowledge organization incorporating taxonomic relationships, (3) uncertainty quantification for practical deployment decisions, and (4) integration with retrieval-augmented generation for improved pattern matching.
This work establishes interpretable VLM-based bioacoustic classification as a viable research direction for human-AI collaborative conservation applications, demonstrating promising preliminary results while acknowledging significant opportunities for improvement.


\small
\bibliographystyle{unsrt}
\bibliography{references}
\normalsize

\end{document}